\documentclass[sigconf]{acmart}

\AtBeginDocument{%
  \providecommand\BibTeX{{%
    \normalfont B\kern-0.5em{\scshape i\kern-0.25em b}\kern-0.8em\TeX}}}
 
\usepackage{pbox}
\usepackage[normalem]{ulem}
\useunder{\uline}{\ul}{}
\usepackage{graphicx}
\usepackage{url}
\usepackage{multirow}
\usepackage{colortbl}
\usepackage{siunitx}
\usepackage{booktabs} % For formal tables
\usepackage{graphicx}
\usepackage{caption}
\usepackage{subcaption}
\usepackage{amsthm}
\usepackage[linesnumbered,ruled,vlined,commentsnumbered]{algorithm2e}
\usepackage{latexsym}
\usepackage{adjustbox}
\usepackage{paralist}
\usepackage{amsmath}
\usepackage{color}
\usepackage{marvosym}
\usepackage{ifsym}
\usepackage{url}            % simple URL typesetting
\usepackage{booktabs}       % professional-quality tables
\usepackage{amsfonts}       % blackboard math symbols
\usepackage{nicefrac}       % compact symbols for 1/2, etc.
\usepackage{microtype}      % microtypography
\usepackage{mathtools}% Loads amsmath
\usepackage{hyperref}      % hyperlinks
\usepackage{amsmath}
\usepackage{amsthm}
\usepackage{amsfonts}
\usepackage{amssymb}
\usepackage{graphicx}
\usepackage{booktabs}
\usepackage{placeins}
\usepackage{multirow}
\usepackage[ruled,linesnumbered]{algorithm2e}
\usepackage{epstopdf}
\usepackage{color}
\usepackage[T1]{fontenc}
\usepackage{lmodern}
\usepackage{amsmath,amsfonts,bm}
\usepackage{enumitem}
\usepackage{xcolor}
\usepackage{wrapfig}
\usepackage{enumitem}
\usepackage{color}
\usepackage{paralist}
\copyrightyear{2020}
\acmYear{2020}
\setcopyright{licensedothergov}\acmConference[JCDL '20]{Proceedings of the ACM/IEEE Joint Conference on Digital Libraries in 2020}{August 1--5, 2020}{Virtual Event, China}
\acmBooktitle{Proceedings of the ACM/IEEE Joint Conference on Digital Libraries in 2020 (JCDL '20), August 1--5, 2020, Virtual Event, China}
\acmPrice{15.00}
\acmDOI{10.1145/3383583.3398541}
\acmISBN{978-1-4503-7585-6/20/06}
\renewcommand\footnotetextcopyrightpermission[1]{} % removes footnote with conference information in first column
\begin{document}
%
% end of the preamble, start of the body of the document source.
%\pagestyle{plain}
%\settopmatter{printfolios=true}
%Your title must be in mixed case, not sentence case. 
% That means all verbs (including short verbs like be, is, using,and go), 
% nouns, adverbs, adjectives should be capitalized, including both words in hyphenated terms, while
% articles, conjunctions, and prepositions are lower case unless they
% directly follow a colon or long dash

\title{Aspect-based Sentiment Analysis of Scientific Reviews}
%
% The "author" command and its associated commands are used to define the authors and their affiliations.
% Of note is the shared affiliation of the first two authors, and the "authornote" and "authornotemark" commands
% used to denote shared contribution to the research.

\author{Souvic Chakraborty}
\affiliation{%
  \institution{Indian Institute of Technology, Kharagpur}
  \state{West bengal}
  \country{India}}
\email{chakra.souvic@gmail.com}

\author{Pawan Goyal}
\affiliation{%
  \institution{Indian Institute of Technology, Kharagpur}
  \state{West bengal}
  \country{India}}
\email{pawang@cse.iitkgp.ac.in}

\author{Animesh Mukherjee}
\affiliation{%
  \institution{Indian Institute of Technology, Kharagpur}
  \state{West bengal}
  \country{India}}
\email{animeshm@cse.iitkgp.ac.in}

% email address must be in roman text type, not monospace or sans serif

%
% The "title" command has an optional parameter, allowing the author to define a "short title" to be used in page headers.
\fancyhead{}
\renewcommand{\shortauthors}{Souvic et al.}

%
% The abstract is a short summary of the work to be  presented in the article.
\begin{abstract}
Scientific papers are complex and understanding the usefulness of these papers requires prior knowledge. Peer reviews are comments on a paper provided by designated experts on that field and hold a substantial amount of information, not only for the editors and chairs to make the final decision, but also to judge the potential impact of the paper. In this paper, we propose to use aspect-based sentiment analysis of scientific reviews to be able to extract useful information, which correlates well with the accept/reject decision.

While working on a dataset of close to 8k reviews from ICLR, one of the top conferences in the field of machine learning, we use an active learning framework to build a training dataset for aspect prediction, which is further used to obtain the aspects and sentiments for the entire dataset. We show that the distribution of aspect-based sentiments obtained from a review is significantly different for accepted and rejected papers. We use the aspect sentiments from these reviews to make an intriguing observation, certain aspects present in a paper and discussed in the review strongly determine the final recommendation. As a second objective, we quantify the extent of disagreement among the reviewers refereeing a paper. We also investigate the extent of disagreement between the reviewers and the chair and find that the inter-reviewer disagreement may have a link to the disagreement with the chair. One of the most interesting observations from this study is that reviews, where the reviewer score and the aspect sentiments extracted from the review text written by the reviewer are consistent, are also more likely to be concurrent with the chair's decision.
\end{abstract}

\maketitle

\thispagestyle{empty}

\section{Introduction}
Peer reviewing is an integral part in identifying important and relevant research in any specific area of science. This has been adopted by major journals and conferences in many areas of science including computer science. However, peer reviewing is manual and it takes a huge amount of time even for an expert in that particular field to review a scientific document. Researchers have spent approximately 63.4 million hours on peer reviews in 2015 alone~\cite{Kovanis2016}. Nevertheless, the process is not foolproof and there have been many recent studies on the inefficacy of the peer-reviewing system and many alternate systems have been proposed. A major conference in the field of computer science did a study in 2014, assigning 10\% of their papers to two different groups of reviewers to analyze the reproducibility and subjectivity in reviews. Disagreement over the accept/reject decision was found for a quarter of those papers even when that conference adopts a three-reviewer majority-vote based algorithm for taking the final decision for any paper~\cite{Langford_2015}.

 \citet{Ragone2013} found that there is very little correlation between peer reviewers' comments and citation counts of these papers. They also report that the peer reviews are not very effective in finding the major flaws in a paper.

These criticisms make studies on peer review dataset very important as it helps to improve science directly by identifying flaws in the scientific process of highlighting best research thus inspiring better feedback mechanisms to draw attention of serious researchers. However, till now, there have been very few studies on this front mainly due to unavailability of peer-review dataset in the public domain.

To address this need, we perform a large scale analysis on the public dataset of peer reviews made available by one of the largest conferences in the area of machine learning -- ICLR\footnote{\url{https://openreview.net/group?id=ICLR.cc}}. ICLR has peer reviews for both the accepted and rejected papers available with rebuttal comments.

We postulate that the reviewers largely follow a well-defined structure while writing reviews identifying the pros and cons of the paper and a peer review corpus is therefore sentiment rich as it is supposed to reflect the qualities of the paper to help the chair/editor understand whether it should be accepted. In particular the reviews would be organized around various aspects and the reviewers would tend to express varying sentiments across these aspects.

In this paper we adopt the reviewing aspects used in the ACL conferences. We manually inspected the reviewing guidelines of various top CS conferences and finally selected the ACL format. This choice is motivated by the level of details that the ACL format brings on board. As per the ACL reviewing guidelines, the eight different aspects that a paper is rated on are -- \textit{appropriateness}, \textit{clarity}, \textit{originality}, \textit{empirical/theoretical soundness}, \textit{meaningful comparison}, \textit{substance}, \textit{impact of dataset/software/ideas} and \textit{recommendation}. 

However, note that our peer review corpora --ICLR 2017, 2018 and 2019 -- are not annotated based on these aspects. Therefore we design an annotation framework to annotate each review with the relevant aspects and the corresponding aspect sentiments present in the review. Following this we perform rigorous data analysis to discover many interesting results illustrated with figures and tables in the later sections. We present the most intuitive explanations of the results in their individual sections as well as in the conclusion. Our specific contributions are enumerated below.
\begin{itemize} 
    \item We collect peer review data for one of the top machine learning conferences -- ICLR. In addition, for every paper we also separately collect various metadata like the authors, the abstracts, rebuttal comments, confidence score of the reviews etc.
    \item We design an annotation framework and manually annotate $\sim 2,500$ review sentences at two levels -- (i) identify and mark the aspects present (if any) in the sentence and (ii) the sentiment associated with each of the identified aspect. We use an active learning framework to choose the sentences to be annotated.
    \item We design a supervised algorithm to label the aspects and the sentiments for the rest of the corpus.
    \item We analyze the machine-labelled data to obtain various interesting insights. The aspect sentiment polarities for the accepted papers are substantially different from those of the rejected papers. For accepted papers, the reviewers express higher positive sentiments in \textit{appropriateness}, \textit{clarity}, \textit{originality}, \textit{empirical/theoretical soundness}, \textit{substance}, \textit{impact of ideas} and \textit{recommendation}. On the other hand, the rejected papers have higher negative sentiments in aspects like \textit{appropriateness}, \textit{clarity}, \textit{empirical/theoretical soundness}, \textit{substance}, \textit{impact of ideas} and \textit{recommendation}. 
    \item We further identify why the chair might differ from the mean recommendation of the reviewers and establish the importance of review texts through that. In particular, we define inter-reviewer disagreement and establish that aspect level disagreements have clear correlation with aggregate level disagreement. Evidently, the chair disagrees more in cases where the correlation between the recommendation score and the corresponding review aspect sentiments are low, i.e., where scores exhibit individual bias on part of the reviewers (which is different from the normative behavior of the reviewers.
    \item Finally, one of the most interesting findings is that sentiments associated with aspects like \textit{clarity}, \textit{empirical/theoretical soundness} and \textit{impact of ideas} strongly determine the intended recommendation in a review. In both the methodologies (linear and non-linear), used to determine the importance of the aspects, we come up with similar results. This might be effective in telling the authors which aspects they should focus on in order to get their papers accepted.
\end{itemize}

\noindent\textbf{Organisation of the paper}: The rest of the paper is structured as follows. In Section~\ref{related_work} we review some of the relevant works in this area. In 
Section~\ref{dataset} we describe the ICLR review dataset that we work on and how we label the data with aspect sentiments. In the next section we perform an exploratory analysis of the aspect sentiment annotated reviews. We study the disagreement of the reviewers in Section~\ref{disagree}. We find how the different aspect sentiments in the review text are correlated to the overall recommendation~\ref{importance}. Finally we conclude in Section~\ref{conclusion}.

\if{0}
t rL reasicvd biewsto  the structure is defined with 8 aspects, namely appropriateness, clarity, originality, empirical/theoretical soundness, substance, impact of dataset, impact of software and recommendation. In absence of ACL review dataset with gold-standard annotation, we design a framework to rate the review sentences on each aspect, detection and aspect sentiment prediction. We train a classifier on this annotated corpus to generate aspect sentiment labels for each sentence for the reviews collected. 

For downstream task like recommendation prediction, we aggregate these sentence sentiments in some way to use for prediction. We also perform various correlation studies on the NIPS dataset to find how the long term citation counts correlate with the review sentiments and recommendation scores.

Our contribution in this paper can be summarized in 3 points:

1. Collection of large dataset with reviews, abstract and other metadata for papers and yearwise citation counts.

2. Performing annotation using an efficient framework and train a classifier to further label the remaining data to be used for analysis.

3. Performing large scale correlation study and establishing the superiority of feature based deep learning architectures over others in task of recommendation prediction.

Finally, we hope that our study will help our fellow researchers to understand the efficacy of the peer-reviewing process and will inspire people to think of ways to improve the same.
\fi

\section{Related Work}\label{related_work}
We divide the related work into two broad segments. The first discusses the techniques and developments in sentiment analysis. The second details the studies so far done on peer-review text.
\subsection{Sentiment prediction}
Detection of sentiment and its classification has been the focus of research in sentiment analysis. Researchers have used unsupervised methods~\cite{Turney2002}, lexicon based methods~\cite{Kaushik2014,Melville2009}, transfer learning~\cite{Yoshida2011,Howard2018}, deep learning architectures~\cite{Zhao2018,Lei2018} etc. to predict the sentiments.

Variants of this main task have also been extensively proposed. Apart from the sentiment prediction of the sentence as a whole, other tasks like aspect-level sentiment prediction~\cite{Huang2018} and target-dependent sentiment prediction ~\cite{Song2019} are also in place. 

For instance, aspect based sentiment analysis has previously been carried out on product reviews~\cite{Song2019,Huang2018}, movie reviews~\cite{Thet2010}, tweets~\cite{Bahrainian2013}, hotel reviews~\cite{al2018} and so on. It has also been observed that the classifiers trained in sentiment prediction task in one domain do not generalize well when applied on a different dataset from a different domain~\cite{Pang2008}. 

The specific task that we solve here does not have an aspect-sentiment annotated peer review dataset and hence we make a crucial contribution by building such a dataset. In addition we use different classifier architectures and perform rigorous experiments on the validation dataset to choose the best among these architectures.

\subsection{Analysis of peer review dataset}

%\tdo{Put one to two line description of PeerRead, MIL, MILAM, our JCDL 2017 work on citation prediction based on review data of JHEP (https://doi.org/10.1109/JCDL.2017.7991572) and our CIKM 2016 work on anomalous reviewer and editor detection (https://dl.acm.org/citation.cfm?doid=2983323.2983675).}
In absence of a public domain peer review dataset with sufficient datapoints, most works on peer review before 2017 have been on private datasets limiting the number of works. In a case study on peer review data of Journal of High Energy Physics,~\citet{Sikdar2016} has identified key features to cluster anomalous editors and reviewers whose works are often not aligned with the goal of the peer review process in general.
~\citet{Sikdar2017} show that specific characteristics of the assigned reviewers influence the long-term citation profile of a paper more than the linguistic features of the review or the author characteristics. Gender bias in peer-review data has been studied in~\cite{helmer17}.~\citet{tomkins17} established that a single blind reviewing system extends disproportionate advantage to the papers authored by famous scientists and scientists from highly reputed institutions.

~\citet{Kang2018} collects and analyzes openly available peer-review data for the first time and provides several baselines defining major tasks. ~\citet{Wang2018} uses state-of-the-art techniques like a Multiple Instance Learning Framework~\cite{Dooly2001,Angelidis2017} with modifications using attention mechanism using the abstract text for sentiment analysis and recommendation score prediction.~\citet{Gao2019} shows that author rebuttals play little role in changing the recommendation scores of a paper while a larger peer pressure plays a greater role for the non-conforming reviewer in the rebuttal phase. %All this work shows major flaws in the peer-review process itself and provides clues that the review text alone contains little information about the long-term citation pattern. We verify these claims in rigorous details in our work via analyzing the aspect sentiment impressions.

\section{Dataset}\label{dataset}

Machine learning conferences like ICLR have adopted an `open review' model making all the comments, author rebuttals and reviews public.~\citet{Kang2018} has open sourced the framework to collect data from ICLR. However, on close observation of the data quality, we decided to collect the data again using our own framework to avoid mixing of some parts of the data like author rebuttals with the reviews which might have introduced specific biases in the dataset. %Some representative examples requiring us to collect the data again are noted in Table~\ref{tab1}. %snapshot of this data sanity check is provided in table no 1.
\if{0}
\begin{table}[t]
\begin{tabular}{llll}
\hline
          & PeerRead & Our data & Data quality \\ \hline
E.g. 1 & wrongtext1     & correcttext1    & why?        \\ %\hline
E.g. 2 & wrongtext2    & correcttext2    & why?        \\ %\hline
E.g. 3 & wrongtext3   & correcttext3    & why?        \\ \hline
\end{tabular}
\caption{\label{tab1} Examples illustrating the need for the collection of the data again.}
\end{table}
\fi
\if{0}Kang et al. also annotated the ICLR reviews from 2017 and 2018 according to ACL 2016 guidelines on 7 aspects and did a correlation study on the same. However, that dataset is not opensourced and can not be used for training.

For NIPS reviews, no aspect-annotated dataset is available. Now, if we perform annotation at document/review level training the classifiers will be very difficult as the number of datapoints will be very low compared to the number of features. So, like Wang and Wan did for their state of the art architecture in sentiment prediction for peer review texts, we ignored the context of the sentences and treated each of the sentences independently. We trained the classifier to detect and extract aspect based sentiments using various architectures. We chose the best architecture for testing purpose.\fi

\subsection{Description of the raw dataset}
ICLR reviews are completely public and all parts of the reviews are readily accessible online. All the comments and author rebuttals are also available. Final score of recommendation is available in a scale of 1-10 for each of the individual reviews. We mined a total of 5,289 ICLR  papers (main conference + workshops); however before 2017 this data is very noisy and incomplete. In fact, prior to 2017, the volume of the data is very less and the quality of the data is also poor as the author rebuttals cannot be separated from the reviews automatically owing to the design of the openreview website. Therefore we use the data for the years 2017, 2018 and 2019 only which makes a total of 3,343 papers. The statistics of the data is noted in Table~\ref{tab:data_dist_iclr_nips}. %On the other hand, NIPS has made public only the after-rebuttal reviews of papers accepted between 2013-2017 on their site. No score associated to each review is provided. There are a total of 3,429 NIPS accepted papers. We collect all these publicly available peer-reviews from the above two sources. In addition, we also collect the author information for each paper as well as some other metadata. %For both the datasets we crawl the citations as in April 2019. For our analysis, for the NIPS dataset, we only use citations for the papers from 2013-16, and for ICLR, we use citations for the papers from 2017.

%Further, for the 1,786 ICLR papers, we collect their citation counts from the Google Scholar database\footnote{We collect citations of only 2017 papers since 2018 papers have very low or no citations at all.}. \todo{PG: Not very clear from the way it is written. Did you collect citations only for 2017 ICLR paper and all the NIPS papers? If so, Why for NIPS 2018 but not for ICLR 2018?}

% \usepackage{graphicx}
\begin{table}[t]
\centering
\resizebox{8cm}{!}{%
\begin{tabular}{l | c c c c}
\hline
        & Oral & Poster & Workshop & Rejected (All tracks) \\ \hline
2017  & 15  &  183  & 48 & 245\\ \hline
2018   & 23  &   313 & 195  &634 \\ \hline
2019  & 24  &   478     & 108 & 1077 \\ \hline

\end{tabular}%
}
\caption{Number of papers in the ICLR dataset from 2017 to 2019. Note that some of the papers may have been withdrawn and we have removed those entries from the calculations hereafter.}
\label{tab:data_dist_iclr_nips}
\end{table}

%We collected the citation information of these 3429 NIPS papers crawling the Scopus database. We collected the citations of 1678 ICLR papers from the google scholar database.
%Our dataset is enriched further with the labels we got using the argument proposition annotation framework published in NAACL-2019. Then we annotate selected sentences from the reviews to train the aspect predictor.
\subsection{Data annotation process}

Peer reviews consist of several technical words and most of the sentences in peer reviews are devoid of any sentiment. Thus, it is difficult to come up with a framework to get enough training data labelled to train the classifier where the resource is limited. Therefore, we use a special type of batch-mode active learning framework to annotate the data.

\subsubsection{Avoiding cold start}
If we start annotating using a random pool of sentences, we may find that most of the sentences are non-informative and devoid of any sentiment. Thus, if we had a mechanism to filter out sentences which are quite likely to have the presence of the aspect sentiments, we may get a more balanced corpus. %\if{0}Researchers have trained and used many classifiers to detect bias for this purpose in other domains but these methods are domain specific and do not work well in cross-domain transfer learning scenarios. In the absence of an annotated dataset, we could not use these pre-trained classifiers for our purpose.\fi

We tackle the problem as follows. We choose a set of 50 sentiment-bearing sentences equally distributed across various aspects (mentioned below) from randomly chosen reviews, and generate sentence embeddings using Google's universal sentence encoder ~\cite{Cer2018}. Then we choose 30 closest sentences in the embedding space in terms of cosine similarity for each of these seed sentences. We repeat the same process once again taking these newly found sentences as seed. Finally, we use the $n$-gram based Jaccard similarity metric to put these sentences into 300 clusters using $K$-means clustering and randomly select 1,500 sentences from these clusters with equal representation from each cluster. We finally get a reduced set of 1,321 sentences after the final label selection using the aggregation methodology discussed in the \textit{Annotation guidelines} section.

\subsubsection{Collecting more data for annotation}

In order to generate more annotated data we use the standard batch selection method used for active learning. Using the already collected batch of sentences, we train a simple random forest classifier. Now we run this classifier on the rest of the sentences to label them. From this set of automatically labelled sentences, we choose 15,000 sentences, 70\% of which are taken from the high entropy zone, i.e., where the classifier is most confused (we take the sentences with entropy larger than 0.5) and the 30\% from the rest of the sentences. One can imagine that for the 70\% high entropy sentences, the confusion of the classifier is so high that these can be literally considered `unannotated' and hence the need for manual annotation. We again use the $n$-gram based similarity as earlier and decompose the sentences into 300 clusters and choose another 1,500 sentences randomly representing all the clusters equally. %We get these 1,500 sentences annotated by our three experts as per the same guidelines outlined in the previous subsection.

\subsubsection{Annotation guidelines}

Each of the above 3,000 sentences are annotated by three experts who are ML/NLP researchers and are quite familiar with the peer-review process in these areas. In our annotation framework, each annotator has to label every sentence with the aspects present in the sentence and the corresponding sentiment that binds to that aspect. For our work, we use the aspect labels used for the ACL 2016\footnote{\url{http://mirror.aclweb.org/acl2016/}} reviews. There are eight aspects, namely, \textit{appropriateness}, \textit{clarity}, \textit{originality}, \textit{empirical/theoretical soundness}, \textit{meaningful comparison}, \textit{substance}, \textit{impact of dataset/ software/ ideas} and \textit{recommendation} in this guideline. The sentiment binding to an aspect could be one of the four labels -- `positive', `negative', `neutral' and `absent'. We ask the annotators to annotate the aspect and the sentiment labels present in each sentence. Note that a particular sentence might not have any aspect (and therefore no sentiments) or multiple aspects with a sentiment binding to each of these aspects. Two annotators work on the annotation at first. In case, two of the annotators agree on a specific label, it is taken for training, else, the third annotator is asked to resolve the disagreement. The third annotator may choose one of the two labels provided by the other two annotators or mark the sentence as ambiguous. We annotated a total of 3,091 sentences and found that the first two annotators agree for 63.1\% of the sentences (1,951). Out of the remaining 1,140 sentences, 502 sentences were discarded as ambiguous by the third annotator, leaving us with a total of 2,589 annotated sentences.  %\todo{PG: May have to provide statistics on this, how many were discarded? This would help to understand how easy the task is for humans.}

%review framework, the annotators are asked to annotate each sentence with 4 labels for each of the 6 aspects. These 4 labels are POSITIVE, NEGATIVE, NEUTRAL and ABSENT and they jointly contain information about the presence and polarity of the sentiments. 

%Batch selection for active learning:
%We train a random forest classifier 

%run a random forest classifier on the remaining unlabelled sentences and based on the entropy of the classifier, chose 15000 sentences for 70 percent of the sentences coming from high-entropy zone (sentences with entropy greater than 0.5) and 30 percent from the rest. We used the same ngram-based clustering for 300 clusters and selection procedure as we did on the last round to select 1500 sentences for further annotation.

After the annotation, we present the statistics for various aspects and sentiments in Table~\ref{tab:annostats}.

\begin{table}[h]
\centering
%\scriptsize
\resizebox{8cm}{!}{%
\begin{tabular}{l|ccc}
\hline
                   Aspects              & Positive & Negative & Neutral \\ \hline
Appropriateness  & 12 & 83 & 11 \\ %\hline
Clarity  & 198 & 359 & 20 \\ %\hline
Originality  & 166 & 65 & 24 \\ %\hline
Empirical/Theoretical Soundness  & 71 & 278 & 64 \\ %\hline
Meaningful Comparison  & 53 & 182 & 25 \\ %\hline
Substance  & 20 & 54 & 2 \\ %\hline
Impact Of Ideas/Results/Dataset  & 171 & 152 & 38 \\ %\hline
Recommendation  & 32 & 43 & 9 \\ \hline
\end{tabular}%
}
\caption{\label{tab:annostats}Distribution of sentiments across various aspects in the annotated 2,589 sentences. Note that, we only present the distributions for the sentences where an aspect is present.}
\end{table}

%\subsection{Citation data}

%We crawl the Scopus database for the NIPS papers and collect citation data for the 2013-16 papers, while we used the Google Scholar database to collect citation data of the accepted ICLR papers. The statistics of the citation data for NIPS is detailed in Table~\ref{tab:citation}. For the ICLR dataset the citation records were mostly available for the 2017 papers at the time of our crawl; the mean and the standard deviation of the citation of the 2017 papers is 108.26 and 143.83 respectively. Both of the datasets have been crawled in April, 2019. \todo{As I mentioned earlier, better to give the date, since this stat may not be valid any more if you are including year 2019.}%~\ref{tab2}.
\iffalse
\begin{table}[h]
\centering
%\resizebox{8cm}{!}{%
\begin{tabular}{c|cc}
\hline
Year & Mean citations ($\mu$) & std. ($\sigma$) \\ \hline
0 & 1.12 & 4.64 \\ %\hline
1 & 4.20 & 12.92 \\ %\hline
2 & 8.18 & 25.59 \\ %\hline
3 & 10.28 & 46.23 \\ %\hline
4 & 13.54 & 80.23 \\ %\hline
5 & 14.64 & 117.02 \\ \hline
\end{tabular}%

\caption{\label{tab:citation} Citation received in the $i^\textrm{th}$ year after publication in the NIPS dataset.}
\end{table}
\fi
\section{Aspect-level sentiment classification}
\if{0}\subsection{Dataset Description:}
Our dataset consists of 2498 review sentences. On average 72.5\% of these sentences contain at least one aspect sentiment. Detailed statistics of the annotate dataset is given in Table%~\ref{tab3}.
\fi

%\subsection{Classifier Architectures: }

We employ several established methods for the aspect-level sentiment classification task. To begin with we use various traditional approaches like Multinomial Na\"ive Bayes (MNB), Random Forest (RF) and SVM with linear (SVM-lin) and rbf (SVM-rbf) kernels.

Next we use a bunch of modern deep learning techniques like BiLSTM with CNN, FFNN fed with (i) Google's universal sentence encoder embeddings (FFNN-uni)~\cite{Cer2018} and (ii) SciBERT embeddings (FFNN-sci)~\cite{Beltagy2019}. Micro-F1 scores (to handle the class imbalance problem) using 10-fold cross-validation for the aspect detection task are reported in Table~\ref{tab4} for each aspect. 
\begin{table}[h]
\centering
%\resizebox{8cm}{!}{%
\begin{tabular}{l|l}
\hline
      Classifiers                                & F1-score \\ \hline
MNB                   & 0.56        \\ %\hline
RF                &    0.62      \\ %\hline
SVM-rbf &    0.62      \\ %\hline
SVM-lin   &  0.60        \\ %\hline
BiLSTM-CNN    &   0.63       \\ %\hline
\rowcolor{green!20}FFNN-uni  &  0.71        \\ %\hline
FFNN-sci     &    0.70      \\ \hline
\end{tabular}%
%}
\caption{\label{tab4}Performance of the classifiers in the aspect detection task.} %\textcolor{red}{So the results are worse than earlier? But it does not change the analysis; I hope we have checked it well?}\textcolor{blue}{The previous scores were reported on micro-average metric. Now they are changed to micro average scores.. hence the decrease. The overall result in the classification task,Table 12 is not changing as the quality have not changed much(some scores are better in the Figures 1 \& 2 as more training data got added to two classes but overall it does not influence the classification scores much) }}
\end{table}

From the results in the table we observe that FFNN-uni outperforms all the other techniques. FFNN-sci is a close competitor; however it takes far more time for training. %\textbf{TODO: But Table 3 shows that RF is doing better?? --> A mistake.. RF value is updated now.}\
We further note the aspect-level sentiment detection performance in Table~\ref{tab5} using the best classifier (FFNN-uni). Since our classes are unbalanced, we have reported the class wise macro averaged F1-score which is most appropriate in such a situation. Note that the scores for Substance and Recommendation are lower than the scores of the other aspects. Lower number of training instances (specifically, for the ``neutral'' sentiment category) may be a possible reason here. Albeit, the classifier is able to pick up useful information in all the cases having better macro-averaged F1-scores than a majority classifier dealing with 4 classes. %\textbf{TODO: Do we need to discuss this? Why Substance and recommendation have a low score (possibly due to small size of the dataset)?--> Added that} 
Some representative examples of aspect sentiments obtained using this model are noted in Table~\ref{tab:examples}. %\textcolor{red}{Are the results macro F1 or micro F1. If it is micro F1, then one can argue that all your good performances can be attributed to sentiments being always classified to ``absent''. In problems where the classes are unbalanced, one has to report macro F1.}\textcolor{violet}{The results are Macro-F1, I wrote the correct values but wrong label, changed now.i thought of reporting the Macro F1 scores also as I have the full classification report with class wise F1 scores, precision and accuracy and support for classes and all on 2 average metric - (micro,macro and weighted). But the macro scores are very high and precision==recall==F1 score~=0.97 mostly for all classes.. }

%We establish here that FFNN with sentence embeddings from Google's pre-trained sentence encoder outperforms all other classifiers on this task based on the F1-scores. FFNN with sentence embeddings from SciBERT is a close competetion but takes far more time in prediction. The F1-scores obtained on the validation set for each classifier is discussed below.

% \usepackage{graphicx}

\begin{table}[h]
\resizebox{8cm}{!}{%
\begin{tabular}{l|ccc}
\hline
                 Aspects                  & Precision & Recall & Macro F1 \\ \hline
Appropriateness                   & 0.83      & 0.94  & 0.86    \\ %\hline
Clarity                            & 0.78     & 0.91   & 0.83    \\ %\hline
Originality                        & 0.69     & 0.79  & 0.72    \\ %\hline
Emp./Theo. Soundness    & 0.62     & 0.70   & 0.65    \\ %\hline
Meaningful Comparison              & 0.78     & 0.88 & 0.81    \\ %\hline
Substance                          & 0.50       & 0.56    & 0.53     \\ %\hline
Impact of Ideas & 0.70     & 0.69  & 0.69 \\ %\hline
Recommendation                     & 0.64     & 0.62   & 0.62     \\ \hline
\end{tabular}
}%
\caption{\label{tab5}Class wise sentiment prediction performance for the best model (FFNN-uni). Since the classes are unbalanced, all values reported here are macro average of the class wise results.}% \textcolor{red}{Please report macro-F1 which is more appropriate.}\textcolor{blue}{two score tables for macro-micro or FFNN-uni and FFNN-sci?}}%. **there should be two score tables.. do asap from same data}

\end{table}

The results show that the classifier fidelity is reasonably well %\textbf{TODO: not sure if we can say that, should we tone down to the classifier performs reasonably well? --> changed good to well.}
and can be reliably used for annotating the full dataset. We therefore run the best classifier on the full dataset and perform a rigorous exploratory analysis in the next section.

\begin{table*}[h]
%\scriptsize
\centering
\resizebox{18cm}{!}{%
\begin{tabular}{l|l|l}
\hline
                        Aspects                         & Polarity & Sentences                                                                                                                                                                                         \\ \hline
\multirow{3}{*}{Appropriateness}                 & Positive & in summary, i think the paper can be accepted for iclr. \\ \cline{2-3} 
                                                 & Neutral   & in the end, it's a useful paper to read, but it's not going to be the highlight of the conference.                                                                                                                              \\ \cline{2-3} 
                                                 & Negative & i find this paper not suitable for iclr.                                                                                                                                            \\ \hline
\multirow{3}{*}{Clarity}                         & Positive & the paper is clear, well organized, well written and easy to follow.                                                                                                                                                             \\ \cline{2-3} 
                                                 & Neutral  & the writing is fairly clear, though many of the charts and tables are hard to decipher without labels                                                                                                                                                    \\ \cline{2-3} 
                                                 & Negative & the paper needs major improvements in terms of clarity .                                                                                                                                          \\ \hline
\multirow{3}{*}{Originality}                     & Positive & the proposed idea is interesting and novel.                                                                                                           \\ \cline{2-3} 
                                                 & Neutral  & this is another simple idea that appears to be effective.                                                                                                                                       \\ \cline{2-3} 
                                                 & Negative & since both approaches have been published before , the novelty seems to be limited.                                                                                                                                \\ \hline
\multirow{3}{*}{Empirical/Theoretical Soundness} & Positive & the experimental results support the theoretical predictions.                                                         \\ \cline{2-3} 
                                                 & Neutral  & lastly, this paper also provides some simulation studies.                                                                                                                                  \\ \cline{2-3} 
                                                 & Negative & i don't think any of the experiments reported actually refute any of the original paper's claim.                                                     \\ \hline
\multirow{3}{*}{Meaningful Comparison}           & Positive & the related work section is complete and well documented.                                    \\ \cline{2-3} 
                                                 & Neutral  & they are the best in terms of precision .                                                      \\ \cline{2-3} 
                                                 & Negative & however, no comparison is clearly made.                                                                                                                                                          \\ \hline
\multirow{3}{*}{Substance}                       & Positive & pros: - the proposed method leads to state of the art results .                                                                                                                                             \\ \cline{2-3} 
& Neutral & pros: - real hardware results are provided. \\ \cline{2-3} 
                                                 & Negative & the experiments do not present any baselines,\\
& & so it is unclear how well the method performs compared to the alternatives.                                                                                   \\ \hline
\multirow{3}{*}{Impact of ideas/results/dataset} & Positive & evaluation on a very large real dataset demonstrates the usefulness of the model for real world tasks.                                                                                                                        \\ \cline{2-3} 
                                                 & Neutral  & the model is tested on a proprietary dataset of real manufacturing product.                                                                                     \\ \cline{2-3} 
                                                 & Negative & the study is restricted to 2-dimensional synthetic datasets.                                                                                                                            \\ \hline
\multirow{2}{*}{Recommendation}                  & Positive & i strongly recommend to accept this paper.                                                                                                                         \\ \cline{2-3} 
& Neutral & after reading their feedback i still believe that novelty is incremental and would like to keep my score. \\ \cline{2-3}
                                                 & Negative & overall, i'm afraid i must recommend that this paper be rejected.                                                                                    \\ \hline
\end{tabular}%
}
\caption{Examples of detected sentiments for different aspects. The sentences chosen are those where the best classifier (FFNN-uni) predicts with a high confidence.}
\label{tab:examples}
\end{table*}

\section{Exploratory data analysis}\label{explore}

Using the automatically annotated full dataset, we investigate the connection between the aspect sentiments of the reviews and the final accept/reject decision of a paper. 

%irst, we baran the classifier on each sentence of the reviews and aggregated the results in paper level using four different metrics for the analysis.

\noindent\textbf{Aggregation of the results}: For the subsequent analysis, we aggregate the results as follows. For a given paper we collect all the review sentences written by all the reviewers. For each aspect, we aggregate the positive and negative sentiment scores separately and normalize each of these scores by the total number of sentences in the reviews for that paper. 

\subsection{Aspect sentiment distribution}

\begin{table}[h]
\resizebox{8cm}{!}{%
\begin{tabular}{l|cc}
\hline
 Conf.         & \#Accepted papers & \#Rejected papers \\ \hline
ICLR 2017 & 198                    & 244                      \\ %\hline
ICLR 2018 & 336                    & 486                       \\ ICLR 2019 & 501                    & 916                       \\ \hline
\end{tabular}
}%
\caption{\label{tab:acc_rej_iclr}Distribution of the accepted and rejected conference papers for the ICLR dataset (not showing withdrawn papers).}
\end{table}

%~\ref{tab:acc_rej_stat}.

\begin{table}[h]
\centering
\resizebox{8cm}{!}{%
\begin{tabular}{l|cc}
\hline
       Parameters                               & Accepted & Rejected \\ \hline
\#Sentences                   &  20.24$\pm$13.34         &    20.77$\pm$12.97     \\ %\hline
Avg. sent. length                   &   18.59$\pm$12.08       & 18.03$\pm$11.77         \\ %\hline
\#Unique words  &    197.14$\pm$93.53      &     195.73$\pm$88.82   \\ \hline
\end{tabular}%
}
\caption{Various statistics for the ICLR review dataset. We see that the reviews for the accepted and rejected papers follow very similar distributions in terms of the number of sentences, average sentence length and number of unique words.}
\label{tab:acc_rej_stat}
\end{table}

In the 2017 -- 2019 block, the total number of accepted ICLR papers is 1,035 and rejected papers is 1,646. The year wise distribution is shown in Table~\ref{tab:acc_rej_iclr}. The number of sentences, the average sentence length and the number of unique words are noted in Table~\ref{tab:acc_rej_stat}. 

\begin{figure}[t]
    \centering
    \includegraphics[width=8cm]{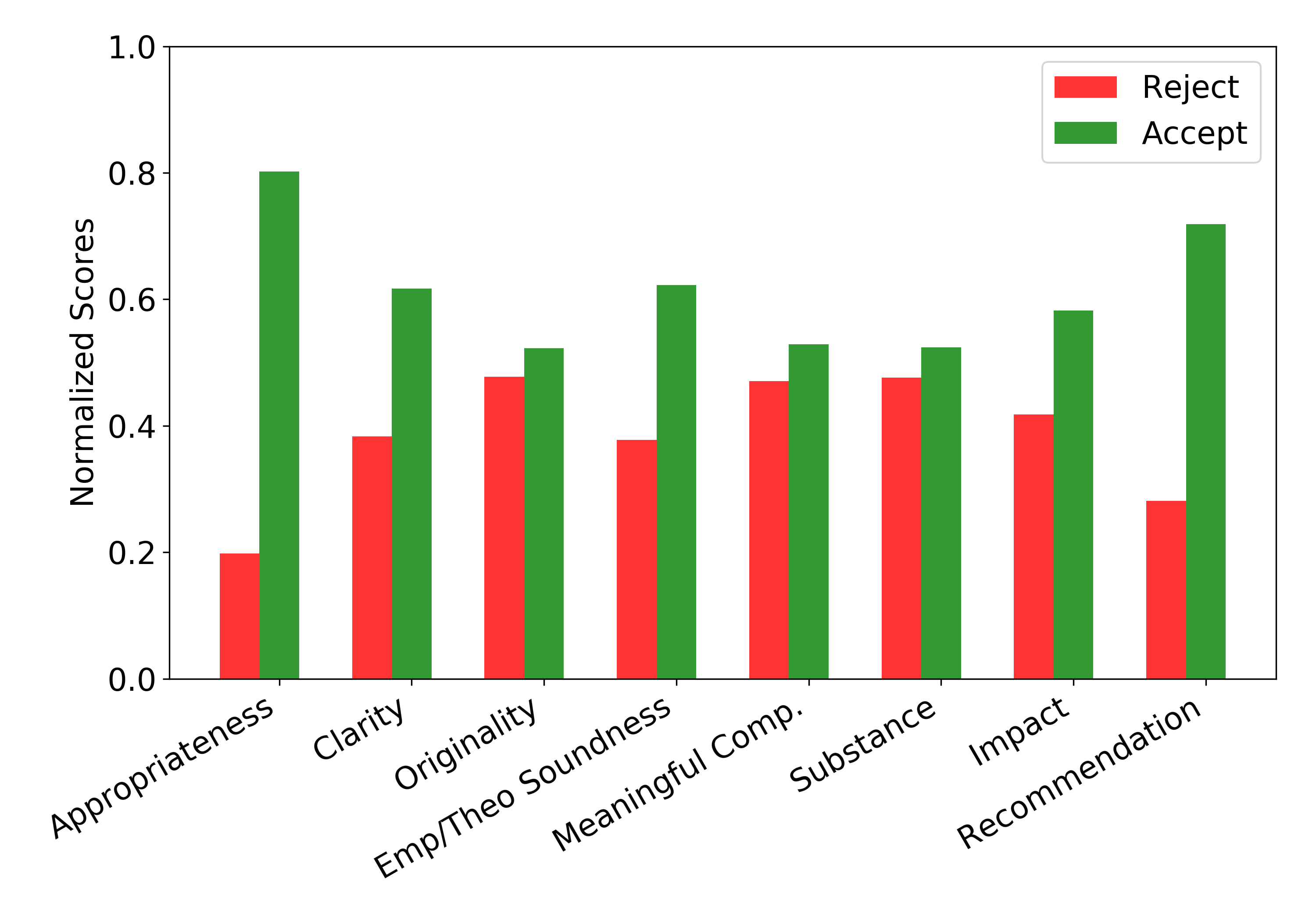}
    \caption{Distribution of positive sentiments binding to different aspects across accepted and rejected papers in the ICLR dataset. Note that the positive sentiment scores for the accepted and rejected papers are normalized across each aspect to add up to 1. Thus, the relative heights across aspects are not comparable.}
    \label{fig:positive_accrej_mean}
\end{figure}

\noindent\textbf{Sentiment distribution}: In Figure~\ref{fig:positive_accrej_mean} we plot the distribution of the positive sentiments binding to different aspects across the accepted and the rejected papers. Note that the aggregation has been done as discussed earlier. After obtaining the mean positive (or negative) scores for all the reviews, we normalize these for each contrasting facet. Thus, while for each aspect, the figure shows the distribution between accept and reject papers, the relative heights of the bars across aspects are not comparable. We use the same strategy across all the figures for easy visualization. The results show some very interesting trends. The accepted papers have a substantially higher positive sentiment for all the aspects -- \textit{appropriateness}, \textit{originality}, \textit{clarity}, \textit{empirical/theoretical soundness}, \textit{meaningful comparison}, \textit{substance}, \textit{impact of ideas} and \textit{recommendation}.

\begin{figure}[t]
    \centering
    \includegraphics[width=8cm]{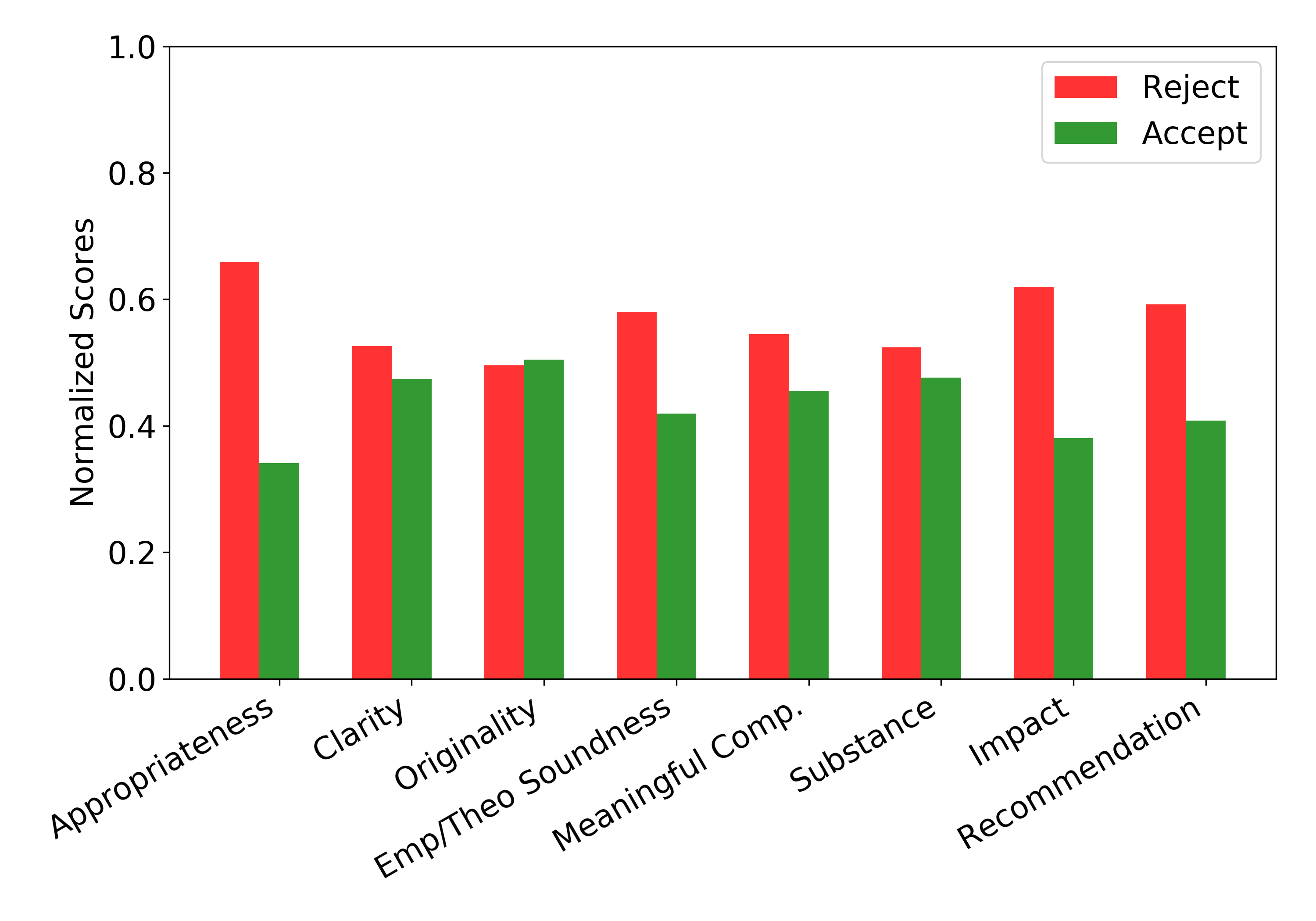}
    \caption{Distribution of negative sentiments binding to different aspects across accepted and rejected papers in the ICLR dataset. Normalization is done similar to Figure \ref{fig:positive_accrej_mean}.}
    \label{fig:negative_accrej_mean}
\end{figure}

In Figure~\ref{fig:negative_accrej_mean} we plot the distribution of the negative sentiments binding to different aspects across the accepted and the rejected papers. The rejected papers have significantly higher negative sentiment for the aspects \textit{appropriateness}, \textit{clarity}, \textit{empirical/theoretical soundness}, \textit{meaningful comparison}, \textit{substance}, \textit{impact of ideas} and \textit{recommendation}. So, apart from the aspect \textit{originality}, where the scores are very close, in all cases, negative sentiments are expressed more for rejected papers on average.

%After pre-processing, we get a total of 753 accepted and 823 rejected papers in ICLR accept-reject dataset as illustrated in Table%~\ref{tab:acc_rej_iclr}.

\subsection{Individual recommendation scores vs the final recommendation decision} We investigate here how the final decision about the paper taken by the chair tallies with the individual reviewer decisions. In order to quantify individual reviewer decision, we consider their recommendation scores of 1-5 to a paper as reject and 6-10 as accept. Next we estimate what percentage of the reviewers disagree with the final recommendation of the chair. If the review scores were completely random this would be 50\%. In the ICLR dataset we find this to be 24.9\%. If the disagreement is calculated based on majority voting then this comes down to 14.8\% which indicates that the chair's decision may significantly vary from the recommendation of the majority of the reviewers. Also, it indicates that the reviewers tend to be more judicious in deciding which half their review scores will fall in between (1-5) and (6-10) bins than assigning the relative scores within each bin if the chair's decision is assumed to be correct. We also explicitly compute the correlation of the recommendation scores with the final recommendation decision. The correlation statistics using three different measures is shown in Table~\ref{tab:corr1}. Once again we observe that the correlation values never go beyond 0.78.

%
%a table showing disagreement stats with length of review..abstract words vs reco prediction... course features
%

\subsection{Why review texts are important when scores are already present?}

As we explored in the previous section, the weighted mean of the scores correlate well with the final recommendation. However, we also find that the final decision matches with the mean score for almost 90\% of the cases. Thus, for the rest of the cases, the chair has to look at the text more thoroughly and make the final decision rather than depending upon the scores given. Also, it will be difficult to identify which papers will fall in that ``90\%'' due to high rate of disagreement among the reviewers and absence of any other good indicator. A possible explanation for this scenario is that there exists disparity between the review written by the reviewers and the scores given by them. We can check the aspect sentiment scores to verify our assumption.

\begin{figure}[h]
    \centering
    \includegraphics[width=8cm]{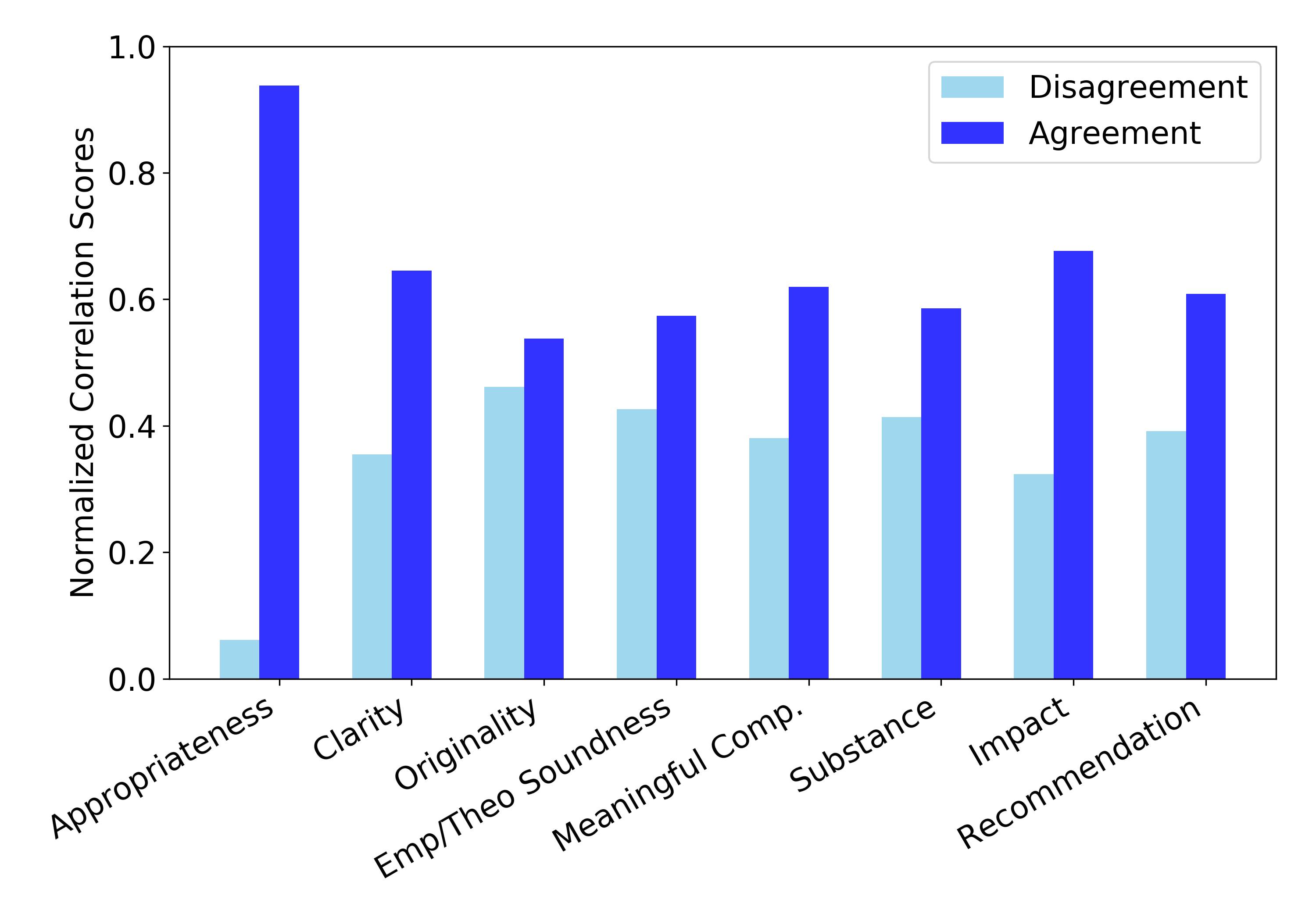}
    \caption{Pearson's correlation coefficient between difference of positive and negative sentiments binding to different aspects across reviews which disagree and agree with the final recommendation, respectively. Note that the sentiment scores for the reviews are normalized across each aspect to add up to 1 so that the relative difference in correlation values can be visualized better. Thus, the relative heights across aspects are not comparable. The $p$-values across these correlation scores are less than 0.001.}
    \label{fig:corr_pearson}
\end{figure}

\begin{figure}[h]
    \centering
    \includegraphics[width=8cm]{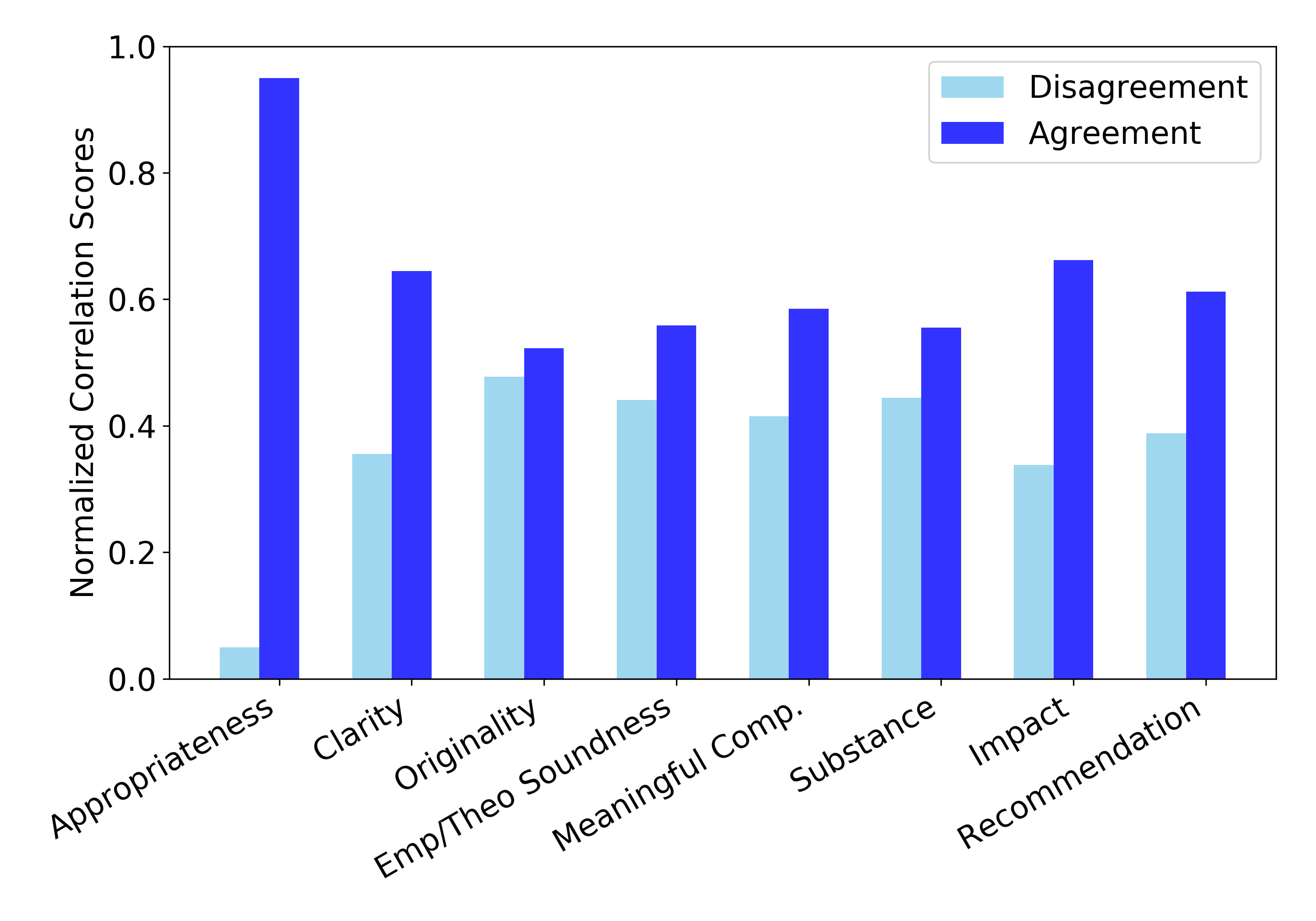}
    \caption{Spearman's rank correlation coefficient between difference of positive and negative sentiments binding to different aspects across reviews which disagree and agree with the final recommendation, respectively. Note that the sentiment scores for the reviews are normalized across each aspect to add up to 1 so that the relative difference in correlation values can be visualized better. Thus, the relative heights across aspects are not comparable. The $p$-values across these correlation scores are less than 0.001.}
    \label{fig:corr_spearman}
\end{figure}

We perform a thorough analysis on the text of the reviews whose score based decision agrees with the chair's decision vs whose score based decision does not, where we define the decision of the reviewer as ``accept'' if the review score is greater than 5 and as ``reject'' otherwise. We run our model to find the aspect sentiment distribution for these reviews. Next we estimate the correlation between each of the aspect sentiment scores with the decision given by that reviewer. We find that the correlation scores are significantly higher in most of the aspects (for other cases the correlation scores are negligible for both cases) for the reviewers whose decisions agree with the chair's decision. The results of this analysis are shown in Fig.~\ref{fig:corr_pearson} and Fig.~\ref{fig:corr_spearman}. We show both the results of rank correlation and correlation by value to strengthen the validity of our results.

Further, we check the average confidence of the reviewers who agree with the chair's decision vs who do not. The mean confidence of the reviewers who agree is 3.88 which is higher than the mean confidence of the reviewers who do not, which is 3.67. We test the statistical significance of this result for these two distributions and find the $t$-statistic to be 6.83 with a $p$-value of 1.07 x $10^{-11}$, thus rejecting the null hypothesis considerably. %\textcolor{red}{Is it $e$ or 10? What is 6.83? Significance only have $p$-values?}\textcolor{blue}{10 , changed.. statistic is the t-statistic giving the ratio of (how different are the mean of the two distributions from each other)/(how different are any sample of any of the two distributions than the whole distribution, i.e. variance measures); whereas the p-value is the probability of rejecting null hypothesis}
Hence, in conclusion, the reviewers who do not agree with the chair's overall decision generally have less confidence and have low correlation between the sentiments expressed in their review text and their review decision. This can be used in future to model the reviewers and to identify the best reviewers as some reviewers clearly are more consistent in text and the scores and their reviews are more likely to help the decision of the chair.

\begin{table}[t]
\resizebox{8cm}{!}{%
\begin{tabular}{l|ccc}  
\hline
                Correlation                       & Pearson's & Spearman's Rank  & Kendall's $\tau$ \\ \hline
    Mean Score   & 0.71   &0.74              & 0.62               \\ %\hline
 Median Score&0.69            &0.71              & 0.63               \\ %\hline
 Majority Voting              &0.67           &0.67              &0.67                \\ 
 Weighted Mean Score             &0.75           &0.78             &0.64               \\ \hline
\end{tabular}%
}
\caption{Correlation of recommendation scores with the final recommendation decision of the chair ($p$-value $<10^{-150}$ in all these cases.)}
\label{tab:corr1}
\end{table}

\section{Disagreement among the reviewers}\label{disagree}

In the previous section we already observed that in around 10\% cases the mean score of the reviewers do not match with the final recommendation. This observation indicates that there are cases of disagreement among the reviewers of a paper on the scores they awarded. In this section, therefore, we study the reviews for the differences of opinion among the reviewers. Intuitively, the papers which are unanimously accepted or rejected must be different from those in which the reviewers disagree upon. However, for the purpose of this investigation, we first need to have a quantitative definition of disagreement.

\subsection{Definition}
We define aggregate disagreement as the standard deviation of the recommendation scores normalized by the maximum standard deviation possible in case of \textit{n} reviewers assigned to every paper. In our case, $n=3$ and the obvious case of highest disagreement arises when the recommendation scores are $1,1,10$ or $10,10,1$. In both the cases, the standard deviation is 4.26 which is used as a constant to normalize the scores.
\subsection{Disagreement distribution}
\begin{figure}[t]
    \centering
    \includegraphics[width=8cm]{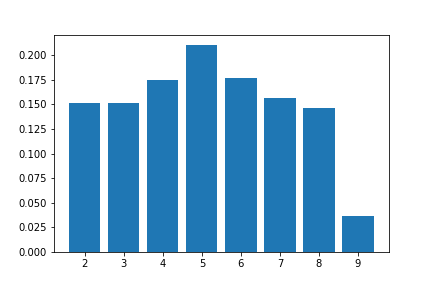}
    \caption{Distribution of average aggregate disagreement values across different bins of average review scores.}
    \label{fig:disagbin}
\end{figure}

In the Figure~\ref{fig:disagbin} we plot the average aggregate disagreement values ($y$-axis) for different bins of average review score that a paper receives ($x$-axis). For instance, $x=2$ represents the set of all papers receiving an average reviewer score in the range [1,2) and the average of the aggregate disagreement values of this set of papers is the corresponding $y$ value. Similarly, $x=3$ denotes [2,3) and so on. We observe that the $y$ values are centered, i.e., the values are non-decreasing till score of 5 and strictly decreasing thereafter. This is quite obvious though as the cases where reviewers are not confident, they are likely to mark a paper in the boundary (score = [5-6)) making the middle part of the curve high in average aggregate disagreement from other reviewers. An alternative possible explanation may be that there is more space to disagree in case when the mean value is near 5 than when the mean value is at the one of the extremes. 

\subsection{Disagreement: Aspect level vs aggregate}
\begin{figure}[t]
    \centering
    \includegraphics[width=8cm]{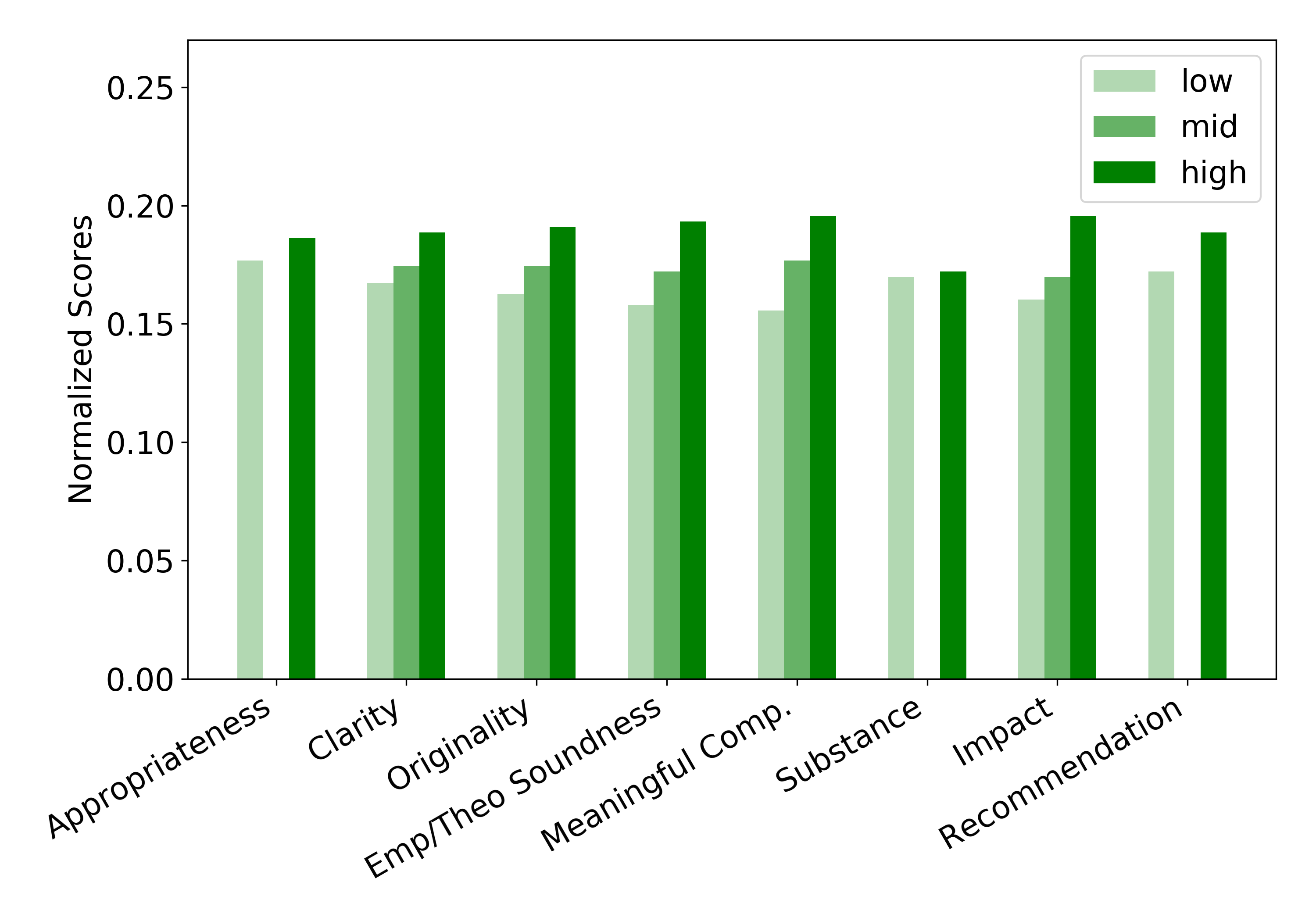}
    \caption{Aggregate vs aspect level disagreements.}
    \label{fig:disag2}
\end{figure}

Here we wish to observe if the disagreement at the aspect level correlates with the aggregate disagreement defined in the previous section. We compute the aspect level disagreement score by taking the standard deviation of the difference of positive and negative sentiment scores for a review. For every paper therefore we obtain an aspect level disagreement score. Next we divide the papers in three bins based on their aspect level disagreement score (low (33$\textrm{rd}$ percentile), high (66$\textrm{th}$ percentile) and mid (between 33$\textrm{rd}$ and 66$\textrm{th}$ percentile)). For the papers in each bin, we compute the average aggregate disagreement score. We plot the results in Fig.~\ref{fig:disag2}. The blank columns signify that there is no paper in that bin. We see that higher disagreement at aspect level on average corresponds to higher disagreement at aggregate level for all the aspects.

\if{0}
\subsection{The effect of the Champion Reviewer}
We wish to study the effect of championing here. Given two papers with a specific mean score, which paper has a higher chance of getting accepted? The one with the higher disagreement or with the lower one? We compute the mean disagreement in case of accepted papers and rejected papers. We found that \fi
\subsection{When will the chair have to intervene?}
In a previous section, we saw that the chair does not go always with the mean decision of the reviewers. We wish to explore if disagreement has some role to play in those cases. We divide the papers in two bins -- (i) bin containing papers where the decision of the chair matches with mean score awarded by the reviewer and (ii) bin containing papers where the decision of the chair does not match. The average aggregate disagreement score for bin (i) is 0.175 which is far lower than bin (ii) where the value is 0.229 (\textasciitilde 30\% higher). Indeed therefore the disagreement among the reviewers is much larger in the latter case needing the chair's intervention.

\if{0}
\subsection{Correlation study:}
We try to find a correlation of these metrics with the final recommendation of the papers and found relatively high correlation for some aspects. We plot the distribution of aspects for positive and negative polarities for accepted and rejected papers in fig. 5 \& 6. We perform similar correlation study on the 2-year citation count of these papers both for the numerical count and the citation bin the paper falls in. We find very weak correlation of recommendation scores with citation profile of the paper as illustrated in Table~\ref{tab:corr2}.

We also explore the correlation of these metrics with individual scores given by the reviewers.

We report the correlation coefficients in the table below. The bar-plots in figures 5-6 shows the sentiment scores on the described four metrics for each of the 6 aspects for accepted and rejected papers. Same is shown for individual accept/reject decision in figures 5-8 and citation counts in figure 9-10.

\begin{table}[]
\centering
\resizebox{8cm}{!}{%
\begin{tabular}{|l|l|l|l|l|l|l|}
\hline
                                       & \multicolumn{2}{c|}{Mean Score}                                                             & \multicolumn{2}{l|}{Median Score}                                                           & \multicolumn{2}{l|}{Majority Voting}                                                     \\ \hline
                                       & Count                               & Bin                                 & Count                               & Bin                                 & Count                            & Bin                                 \\ \hline
Pearson correlation co-efficient       & 0.297                                        & 0.307                                        & 0.290                                        & 0.290                                        & 0.132                                     & 0.207                                        \\ \hline
Spearman rank correlation co-eff. & 0.321                                        & 0.260                                        & 0.321                                        & 0.246                                        & 0.202                                     & 0.181                                        \\ \hline
Tao-Kendall co-efficient               & 0.248                                        & 0.224                                        & 0.248                                        & 0.226                                        & 0.166                                     & 0.179                                        \\ \hline
Max p value in the column              & 4.13x10\textsuperscript{-5} & 3.70x10\textsuperscript{-4} & 6.78x10\textsuperscript{-5} & 8.20x10\textsuperscript{-4} & 7x10\textsuperscript{-2} & 1.45x10\textsuperscript{-2} \\ \hline
\end{tabular}%
}
\caption{\label{tab:corr2}Correlation of recommendation scores with 2-year citation count}
\end{table}
\fi
\section{Determining the importance of the aspects in recommendation}\label{importance}

We conduct two different studies here to understand which features contribute more to the final recommendation. First we observe the correlation of the aspect sentiments of the different aspects and the final recommendation. Second we develop a full-fledged deep neural model that uses the aspect sentiments to predict the final recommendation. The aspect sentiment features used in the model are then ranked to identify the most predictive aspects.

%we use our classifier described in the previous sections to find the correlation score of the aspect sentiments with the final recommendation. Then in the second method, we use a deep learning classifier to predict the final recommendation of each review in order to finally perform a feature-dropout ablation study.

\subsection{The correlation study}\label{corr_study}
We take the difference of average positive sentiments and average negative sentiments for all sentences in the review as the final sentiment scores at aspect level for each of the aspect. We then compute the Pearson correlation coefficients of the aspect level sentiment values and the final recommendation. The result is shown in the Table~\ref{tab:feat_imp}. We can see that the aspects \textit{clarity}, \textit{empirical/ theoretical soundness} and \textit{impact of ideas} have higher correlation values than the other features.
\begin{table}
\centering
\resizebox{8cm}{!}{%
\begin{tabular}{l|c}
\hline
       Aspects                                                           & Pearson Correlation Co-eff \\ \hline
Appropriateness                                             &0.066                                         \\ %\hline
\rowcolor{green!10}Clarity                             & 0.212                                         \\ %\hline
Originality                                & 0.028                                         \\ %\hline
\rowcolor{green!20}Empirical/theoretical soundness                                &0.267                                             \\ %\hline
Meaningful Comparison                              &  0.101                                        \\ %\hline
Substance                                 &0.010                                          \\ %\hline
\rowcolor{green!25}Impact of Ideas                                 &0.273                                           \\ %\hline
Recommendation                              & 0.074                                         \\ \hline
\end{tabular}%
}
\caption{Correlation of the aspect sentiments with the final recommendation.}
\label{tab:feat_imp}
\end{table}

\subsection{Deep neural model}\label{deep_study}

For every individual review for a paper, we attempt to predict the intended decision as to whether that review indicates acceptance or rejection of the paper directly from the review text. Since the data is already small, we integrate the reviews of 2017, 2018 and 2019 together for the prediction task. This makes a total of 8151 reviews.

\noindent{\bf Details of the review text}: Each review in the dataset consists of 20.57 sentences on average and each sentence contains 18.37 words on average. The recommendation scores range from 1 -- 10, 10 being the best. The distribution of the recommendation scores is shown in Table~\ref{tab:iclr_rec_count}.
\begin{table}[h]
\centering
\resizebox{8cm}{!}{%
\begin{tabular}{l|l|l|l|l|l|l|l|l|l|l}
\hline
Scores & 1  & 2  & 3   & 4   & 5   & 6    & 7    & 8   & 9  & 10 \\ \hline
Count  & 16 & 146 & 651 & 1528 & 1645 & 1874 & 1619 & 513 & 146 & 13  \\ \hline
\end{tabular}%
}
\caption{Distribution of recommendation scores in the ICLR dataset.}
\label{tab:iclr_rec_count}
\end{table}
%\subsection{Dropout analysis}
%Using all the observations in the previous section, we propose a downstream task here to observe which feature contributes more to the prediction.  

\if 0
\begin{table}[h]
\centering
\resizebox{8cm}{!}{%
\begin{tabular}{l|l|l|l|l|l|l|l|l|l|l}
\hline
Scores & 1  & 2  & 3   & 4   & 5   & 6    & 7    & 8   & 9  & 10 \\ \hline
Count  & 16 & 146 & 651 & 1528 & 1645 & 1874 & 1619 & 513 & 146 & 13  \\ \hline
\end{tabular}%
}
\caption{Distribution of recommendation scores in the ICLR dataset.}
\label{tab:iclr_rec_count}
\end{table}\fi

\noindent\textbf{Prediction classes}: The review scores 1 -- 10 are translated to accept/reject indicators for the 2-class classification task. For a particular review, we translate the scores 1 -- 5 to a reject decision and the scores 6 -- 10 as accept decision. Thus all the 8151 reviews get tagged with one of the accept/ reject labels in this way. The distribution of the number of reviews in each class is noted in Table~\ref{tab:two-class}.

\begin{table}[h]
\resizebox{8cm}{!}{%
\begin{tabular}{l|cc}
\hline
    Conference      & \#Accepted ($>5$) & \#Rejected ($\le 5$)  \\ \hline
ICLR 2017 & 776                    &   573                    \\ %\hline
ICLR 2018 & 1242                    & 1234                     \\

ICLR 2019 & 2147                    & 2179                     \\ \hline
\end{tabular}%
}
\caption{\label{tab:two-class}Distribution of the reviews for the 2-class classification task.}

\end{table}

\if 0
\subsection{Description of the dataset}
For this task, we collected 4961 reviews with gold-standard annotation set collected from openreview.com . We used general pre-processing techniques like stopwords removal, tokenization, Casefolding etc. to clean the dataset.

Each review in the dataset consists of 19.91 sentences on average and each sentence contains 19.49 words on average. The recommendation scores range from 1-10 , 10 being the best. The distribution of the recommendation scores are shown in Table~\ref{tab:iclr_rec_count}. The distribution of sentence lengths and number of sentences used in the reviews are shown in table no 5. 
% \usepackage{graphicx}
\fi

\noindent\textbf{Model description}: We devise our model to show the usefulness of the aspect sentiments in predicting the recommendation intended in a review and to compute the effect of dropout of individual features in the model to understand the hierarchy of importance of the aspects. For this we build aspect sentiment features as follows.

\noindent\textbf{Aspect sentiment features}: Recall that we have eight aspects. For each aspect we have three sentiment categories -- `positive', `negative' and `neutral'. Therefore we construct a $8\times 3 = 24$ dimensional vector. For each of the eight aspects we keep three cells. We aggregate the sentiments in each cell using the same aggregation method discussed earlier (see section~\ref{explore}) except that we perform $Z$-score normalization on each of the 24 features for faster training.

\noindent\textbf{Prediction}: We pass the 24 dimensional aspect sentiment vector through one dense layer with ReLU activation followed by another dense layer and sigmoid activation function to obtain the final classification result.

\noindent\textbf{Hyperparameters}: We use ADAM for optimization and cross entropy as the loss function with default TensorFlow hyperparameters. As an exercise, we try using different number of nodes in the dense layer, different dropout rates and number of hidden layers and choose the architecture performing the best based on 10-fold cross validation accuracy. We used 64 nodes with a dropout rate of 0.8 in all dense layers.

\noindent\textbf{Results}: We obtain 65.3\% accuracy after 10-fold cross validation in this task making it apparent that the aspect sentiment features hold useful information about the final recommendation. We further investigate the importance of the sentiment features in our task. We drop one aspect (3 sentiment features, positive, negative and neutral) each time and observe the \% drop in the accuracy in Table~\ref{tab:feat_ablation}. While the method used in the last section (i.e., section~\ref{corr_study}) studies linear correlation, this one is a non-linear method.  We note that the sentiments binding to the aspects \textit{empirical/theoretical soundness}, \textit{impact of ideas} and \textit{clarity} are the most effective ones in predicting the intended recommendation in a review. This again confirms the results of the previous section. We believe that this result is particularly important because it acts as a guideline for the authors as to what particular aspects they should focus on while submitting their papers. 

\if{0}
\noindent\textbf{Discussion}: Note that as we have pointed out in the introduction, our main motive in doing this task is not to outperform the state-of-the-art but to show the utility of the aspect sentiments. Our approach is conceptually simple, explainable (see the feature ablation study in the next section) and computationally very fast. Works that are similar in spirit to this task like MIL~\cite{Angelidis2017} and MILAM~\cite{Wang2018} (albeit in a slightly different settings) achieve a marginally higher accuracy but are significantly more complex, highly compute intensive, use features from the paper (abstracts) and the results obtained are hard to explain.
\fi

%Asat we show in table no. 4, our classifier achieves comparable performance with MILAM using additional information through the course features, which is extracted from the same review text and abstract. While DAN uses very simple architecture and trains easily, it is able to beat other complex architectures that uses contextual information. The embedding space is huge and finding efficient bi-gram formation logic is difficult due to unavailibility of huge training data and comparably much larger number of bigrams in the traing set. This might explain the degradation observed in case of other complex deep learning architectures.

%We also perform a feature ablation study to understand the contribution of the aspect sentiment features and course features.
%,,,,,,,
\begin{table}
\centering
\resizebox{8cm}{!}{%
\begin{tabular}{l|c}
\hline
       Aspects                                                           & \% difference in accuracy \\ \hline
                                          Accuracy (all aspects)     &   0.0                                       \\ %\hline
Appropriateness                                             &1.11                                          \\ %\hline
\rowcolor{green!10}Clarity                             & 1.71                                         \\ %\hline
Originality                                & 0.88                                         \\ %\hline
\rowcolor{green!20}Empirical/theoretical soundness                                &3.84                                          \\ %\hline
Meaningful Comparison                              &  1.22                                        \\ %\hline
Substance                                 &0.32                                          \\ %\hline
\rowcolor{green!15}Impact of Ideas                                 &3.15                                          \\ %\hline
Recommendation                              & 0.11                                         \\ \hline
\end{tabular}%
}
\caption{Feature importance study: \% drop in classifier performance while dropping each aspect sentiment feature.}
\label{tab:feat_ablation}
\end{table}

\section{Conclusion}\label{conclusion}
In this paper we presented a detailed analysis of the latent aspect sentiments in the peer review text of the papers submitted to/published in one of the top-tier machine learning conferences -- ICLR. In particular we made the following contributions.
\begin{itemize}
    \item We collected the whole ICLR review corpus for the years 2017-19 with all metadata like author names, rebuttal comments, accept/reject/withdrawn information, etc.
    \item We annotated around 2,500 review sentences with the aspects present in them and the associated sentiments binding to those aspects. We used an
    active learning framework to annotate the data in steps. We do not know of any such data that already exists and stress that this is an important contribution of our work. We plan to release this dataset upon acceptance.
    \item We built aspect sentiment detection models that are trained using the above data and obtain a good performance. This allows us to confidently label the rest of the data with appropriate aspects and sentiments. 
    \item We perform an exploratory analysis on this full dataset and investigate how the aspect sentiment distributions vary between accepted and rejected papers. We also establish that the reviewers whose review scores/ recommendations are contradicted by the chair's decision, generally have low correlation between the raw scores and the aspect sentiments coming from text of their review. Thus, the text contains important cues in these cases to make the final decision which cannot be made from their review scores.
    \item We define inter-reviewer disagreement and perform multiple analysis to establish the connection between aspect level disagreements with aggregate level disagreement empirically.
    \item We observe very clearly through multiple experiments that certain aspects are more relevant in the prediction of the final recommendation. These aspects can act as guidelines for authors telling them what to focus more on in order to get their papers accepted.
    
\end{itemize}

In future we would like to extend this work in multiple directions. One immediate task would be to integrate aspect sentiments in the modern future citation count prediction algorithms and observe if we get additional benefits. Another direction would be to study author specific aspect sentiments, i.e., for a given author can we characterise him/her based on the aspect sentiments present in the reviews of the papers authored by him/her? For instance, are certain aspect sentiments more pronounced for successful/high impact authors? A third line of investigation would be the temporal change in aspect sentiments in the reviews received by an author. Does this change provide early indication of the growth/decline of an author in productivity/citation impact?  

\if{0}
In this paper we explore the review dataset for top tier conferences where the data is publicly available. We annotate the review sentences and label the whole dataset using our classifier establishing high correlation of the final recommendation with the aspects. We also use the classifier to get sentiment labels for all the reviews and establish correlation of these aspect sentiments with long term citation of these papers.

\begin{acks}
We are grateful to IIT Kharagpur for the computing resources that we have used. Additionally, we are grateful to Tata Consultancy services for providing the necessary funds to carry out this research.
\end{acks}
\fi

%
% The acknowledgments section is defined using the "acks" environment (and NOT an unnumbered section). This ensures
% the proper identification of the section in the article metadata, and the consistent spelling of the heading.

%
% The next two lines define the bibliography style to be used, and the bibliography file.
\bibliographystyle{ACM-Reference-Format}
\bibliography{acm}
\end{document}